 \documentclass[final,5p,times,twocolumn,authoryear]{elsarticle}

\usepackage{amsmath}
\usepackage{amssymb}
\usepackage{amsthm}

\usepackage{float}

\usepackage{color}
\definecolor{linkcolor}{rgb}{0,0,0.66} 
\usepackage[pdftex,colorlinks=true, pdfstartview=FitV, linkcolor= linkcolor, citecolor= linkcolor, urlcolor= linkcolor, hyperindex=true,hyperfigures=true]{hyperref} 

\usepackage{bm, mathtools}

\renewcommand{\cite}{\citep}

\renewcommand{\vec}{\bm}

\begin{document}

\begin{frontmatter}



\title{Power Gradient Descent}

\author[label1,label2]{Marco Baiesi}
\address[label1]{Department of Physics and Astronomy, University of Padova, Via Marzolo 8, I-35131 Padova, Italy}
\address[label2]{INFN, Sezione di Padova, Via Marzolo 8, I-35131 Padova, Italy}
\ead{baiesi@pd.infn.it}


\begin{abstract}
  The development of machine learning is promoting the search for fast and stable minimization algorithms. To this end, we suggest a change in the current gradient descent methods that should speed up the motion in flat regions and slow it down in steep directions of the function to minimize. It is based on a ``power gradient'', in which each component of the gradient is replaced by its versus-preserving $H$-th power, with $0<H<1$.   We test three modern gradient descent methods fed by such variant and by standard gradients, finding the new version to achieve significantly better performances for the Nesterov accelerated gradient and AMSGrad. We also propose an effective new take on the ADAM algorithm, which includes power gradients with varying $H$.

\end{abstract}



\begin{keyword}



Optimization \sep Gradient Descent 
  
\end{keyword}

\end{frontmatter}


\section{Introduction}
\label{sec:in}

At the core of most modern machine learning tools, and in general in a huge variety of algorithms, one finds gradient descent (GD) techniques~\cite{meh19}. In neural networks, usually the minimization of a complex cost function converges via some GD method to one of its local minima.
Momentum was introduced in some algorithms~\cite{nes83,qia99,goh17}
so that the dynamics gets progressively more speed in flat directions of the cost function, where otherwise the GD would get stuck for a long time~\cite{sut86}. 
Recent methods as, for instance, RMSprop~\cite{tie12}, ADAM~\cite{kin15} and AMSGrad~\cite{red18}, utilize a running average of the square of the gradient to rescale the norm of the step size, thus stabilizing the steps in the optimization.

We introduce a simple modification that can improve the convergence speed and stability of all GD versions.
Assuming that ``small'' and ``large'' are with respect to quantities of order of one, a power $x^H$ with exponent $0<H<1$ of a quantity $x>0$ results larger than $x$ for small $x$'s, and reduced with respect to $x$ if this is large. Based on this simple observation, to boost the speed of GD in regions of shallow gradients and at the same time stabilize their step size in regions with large gradients, we propose to replace each component of the gradient by its versus-preserving $H$-th power.

\begin{figure}[!b]
  \centering
  \includegraphics[width=7cm]{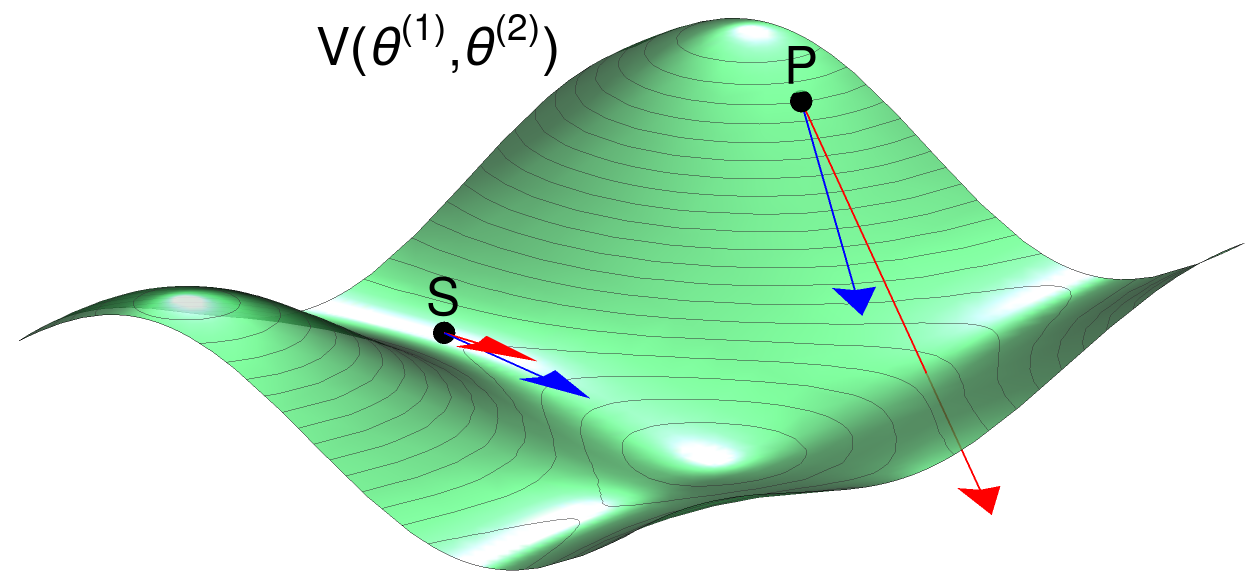}
  \caption{Sketch of a cost function $V$ and of the gradient (red arrows) and corresponding power gradient (blue arrows) at a point within a steep slope (P) and at a point in a shallow region near a saddle (S).}
  \label{fig:sk}
\end{figure}

\begin{figure*}[!tb]
  \centering
  \includegraphics[width=15cm]{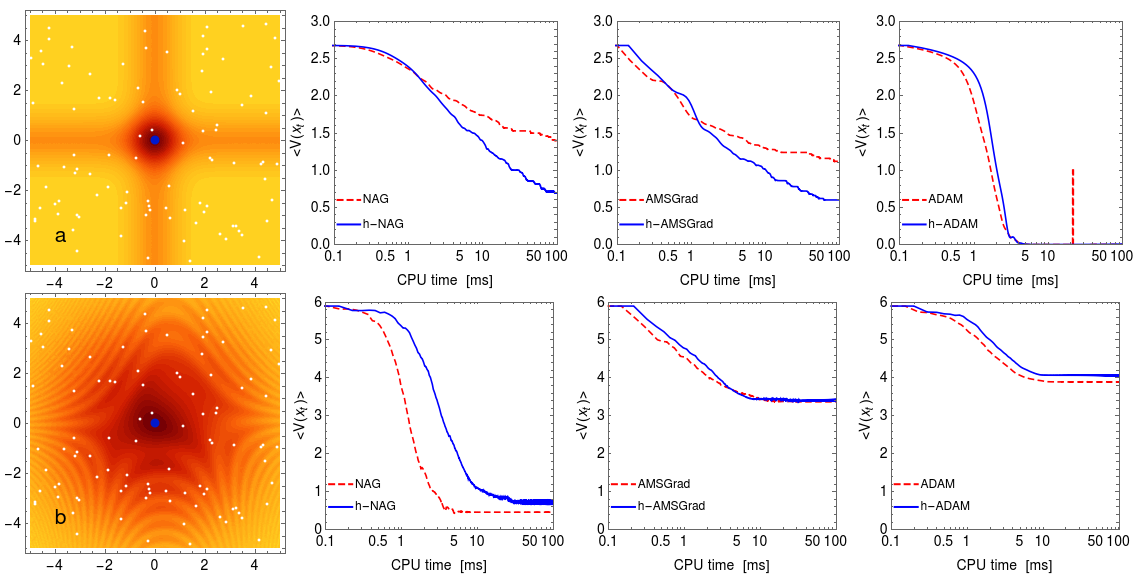}
  \caption{Row (a) for $V_1$ and (b) for $V_2$ show their contour plot in the chosen domain $D$ (white dots at random starting points, and blue dot at the absolute minimum) and, for each simulated GD method, the mean value of the cost function as a function of the mean time of steps in each minimization, in CPU time. Curves are for learning rates leading to best performances: $\eta=10^{-2}$ for NAG variants and $10^{-1}$ for the rest.
  }
  \label{fig:gr1}
\end{figure*}

More precisely, for a cost function $V(\vec \theta)$ that depends on the variables $\vec \theta =(\theta^{(1)},\theta^{(2)},\ldots,\theta^{(N)})$, every component of the gradient is modified as
\begin{equation}
  g^{(i)}(\vec \theta) = \frac{\partial V(\vec \theta)}{\partial \theta^{(i)}}\quad \to\quad
  h^{(i)}(\vec \theta) = {\textrm{sign}}\left(g^{(i)}\right) \left|g^{(i)}\right|^H
\end{equation}
This ``power gradient'' $\vec h(\vec \theta)$ preserves the direction of each component and rescales its magnitude by the $H$-th power. From now on we mostly focus on the $H=1/2$ case, i.e.~we take square roots $|g^{(i)}|^H=\sqrt{|g^{(i)}|}$ if not otherwise stated.
The goal is obtaining the effects sketched in Fig.~\ref{fig:sk}, where normal gradients are represented by red arrows and power gradients by blue arrows. A large gradient (point P) leads to a smaller power gradient, while regions with small gradient (saddle with point S) generate power gradients with larger magnitude. Thus, any GD method adopting the power gradient should be more stable in steep gradient areas and, at the same time, less stagnant in flat regions of the cost function.

We will denote any power gradient variant with the notation ``h-'', for example h-ADAM denotes the ADAM algorithm with $g^{(i)}$ replaced by $h^{(i)}$.

\section{Results}
\label{sec:res}

The step of a vanilla h-GD is,
\[
\theta_{t+1} = \theta_t - \eta\, h(\vec \theta_t)
\]
where we have simplified the notation by removing the component index $(i)$, $t$ is the index of the current iteration, and $\eta$ is the learning rate.
By studying this dynamics in a simple one dimensional quadratic cost function $V(\theta) = \kappa \theta^2/2$, we find the first interesting feature of the h-GD: it does not converge to the minimum!
Even for small $\eta$ values, one easily verifies that the algorithm converges (always) to a 2-state orbit where it keeps bouncing between $-\tilde \theta$ and $+\tilde \theta$, with $\tilde \theta=\eta^2 \kappa/4$.
Hence, a simple new way for testing the convergence is to check whether the sum
$|h(\theta_t) + h(\theta_{t-1})|$ is close to zero, as it should if the two power gradients are opposite to each other.
One may then estimate $V(\overline \theta_t)$ in the middle point $\overline \theta_t = (\theta_t + \theta_{t-1})/2$ of the 2-state orbit.
With these tricks, the h-GD finds the minimum of the quadratic function while remaining stable, i.e.~there is no upper threshold for the learning rate.
In realistic situations there are of course deviations from quadratic $V$'s around its local minima and, heuristically, we may anticipate that $\eta$ needs to be small enough for keeping $\vec \tilde \theta$ within a local quadratic well of $V(\vec \theta)$. However, this argument applies to the basic h-GD algorithm, which, similarly to the basic GD, is less efficient than newer methods.

With more complex GD methods, including momentum and second moment rescaling, we have found that the 2-state orbits are not very relevant and that minimization can be monitored on a single last state $\vec \theta_t$ of a run. Nevertheless, it is important to be aware of the feature described above in the analysis of GD methods embodying power gradients.

\begin{figure*}[!tbh]
  \centering
  \includegraphics[width=14cm]{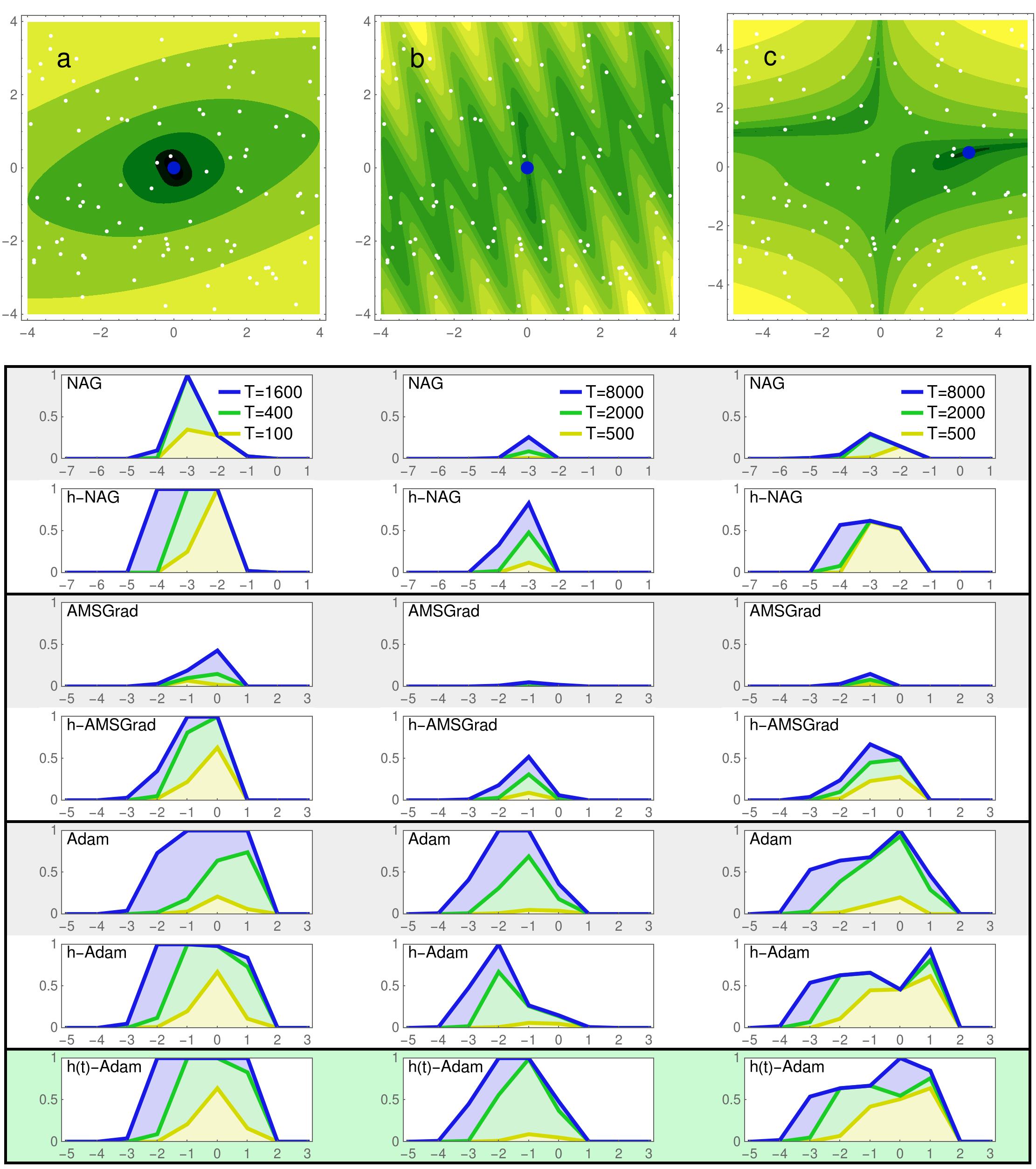}
  \caption{Each column is for a given cost function ((a) $V_3$, (b) $V_4$, (c) $V_{\rm Beale}$) and contains its contour plot in the chosen domain $D$ (contours in log scale, each color shade representing a decade; white dots at random starting points, and blue dot at the absolute minimum) and, for each simulated GD method, the fraction of trajectories converging within a time $T$ (see legends) as a function of the log of the learning rate ($\log_{10} \eta$).
    Rows with gray background refer to traditional methods, white background to the respective power gradient variants with $H=1/2$, and the last row to the h(t)-ADAM implementation.
  }
  \label{fig:gr}
\end{figure*}

The behavior of standard GD techniques and of the new h-GD versions is illustrated with some two-dimensional cost functions $V(\vec \theta) = V(x,y)$.
As representative of methods with momentum, we consider the Nesterov accelerated gradient (NAG)~\cite{nes83}.
To represent modern second moment methods, we study both the AMSGrad~\cite{red18} and ADAM~\cite{kin15} algorithms. All details of their modified versions are reported in \ref{sec:ap}.

We collect the statistics of minimization trajectories starting from $n=100$ initial points spread randomly in a domain $D$, where cost functions have the lowest minimum $V(\vec \theta^{\min}) = 0$ at $\vec \theta^{\min}\in D$.
Hence, to test the convergence in this simple setup we require $V(\vec \theta_t)<10^{-4}$.

We start by studying the convergence speed for bounded cost functions 
\begin{align}
  V_1(x,y) & = -e^{-x^2-y^2}-e^{-x^2}-e^{-y^2}+3 \\
  V_2(x,y) & = 11.5 - 10 e^{\frac{1}{25} \left(-x^2-y^2\right)}-e^{-x^2-y^2}\nonumber\\
  &\quad -0.5 \cos \left(- y x^2 - y^2 + x y + x + 2 y\right)
\end{align}
shown respectively in Fig.~\ref{fig:gr1}(a) and Fig.~\ref{fig:gr1}(b).
The function $V_1$ has a single minimum at $\vec \theta_1^{\min} = (0,0)$ and wide plateaus that hinder the convergence of points initially too far from $\vec \theta_1^{\min}$. This picture is a caricature of neural networks with improperly normalized inputs, leading to tiny gradients because the nonlinear functions, say ReLU functions, are evaluated deep in their flat region.
The function $V_2$ instead represents a complex landscape with ripples and several secondary minima in addition to the absolute minimum $\vec \theta_2^{\min} = (0,0)$.

In Fig.~\ref{fig:gr1}(a) there are the values of $V_1(\vec \theta_t)$ averaged over the $n$ minimizations, as a function of the average CPU time per run; each panel refers to an algorithm in the standard and in the new version.
In this example, the h-NAG is faster than NAG and h-AMSGrad is faster than AMSGrad. This means that the slightly longer code and CPU time per step is compensated by a better overall convergence of the new versions. However, ADAM does not get any improvements from the power gradient, in this example.

In Fig.~\ref{fig:gr1}(b) we show the same plots for the averaged $V_2(\vec \theta_t)$. In this case, introducing the power gradients it is at best not an advantage. The power gradient here does not help in overtaking the ripples of $V_2$. Note that in this example NAG shines as it inertially wins against the little barriers, while ADAM and AMSGrad are too effective in adapting to the local slopes and do not find easily the global minimum.

To challenge further the GD methods, next we test the minimization of unbounded cost functions,
\begin{align}
  V_3(x,y) & = \frac{10}{27} (2 y-x)^4+\frac{10}{9} (2 x+y)^2+5 y^4 \\
  V_4(x,y) & = \log \left((3 x+y)^2+1\right)\nonumber\\
  &\quad +\cosh \left(4 \sin \left(\frac{\pi}{2}   (3 x+y)\right)+x-3 y\right)-1 \\
  V_{\rm Beale}(x,y) & = \left(x y^3-x+2.625\right)^2\nonumber\\
  &\quad +\left(x y^2-x+2.25\right)^2+(x y-x+1.5)^2
\end{align}
where the latter is Beale's test function (see their level plots in Fig.~\ref{fig:gr}). These functions are ranked by increasing complexity: $V_3$ has a single minimum with a simple power-law divergence; $V_4$ has an exponential divergence forming steep walls surrounding a wavy flat ravine, i.e.~a structure that challenges inertial dynamics and stability; finally, the Beale's test function contains narrow ravines, a shallow central region, and global minimum in $\theta_{\rm Beale}^{\min}=(3, 0.5)$ plus a secondary minimum.
These functions contain a mixture of flat and steep directions, which, as we conjectured, is a landscape suitable for h-GD methods.

The quality of an algorithm is now measured by the fraction $f_D(\eta,T)$ of trajectories in the domain $D$ that reached convergence within a time $t\le T$ when the learning rate is $\eta$.
We monitor the behavior of $f_D(\eta,T)$ vs (integer values of) $\log_{10} \eta$. This assigns importance also to the stability of the methods and follows the standard procedure of exploring the performances of algorithms for $\eta$ spanning several orders of magnitude~\cite{meh19}.

In Fig.~\ref{fig:gr} we see that,
in all examples, h-NAG is better than NAG and h-AMSGrad is better than AMSGrad, corroborating the hypothesis that the power gradient is able to speed up the convergence while stabilizing the algorithm. Indeed, both the full curves for $f(\eta,T)$ and their best case scenario at an optimal $\log_{10}\eta$ in general are significantly higher for the h-GD methods than for the GD ones, i.e.~h-GD methods converge faster and eventually with more trajectories in the domain $D$.

For ADAM vs h-ADAM the comparison continues to be not clearly in favor of any of the two versions. Quite likely, the structure of the ADAM algorithm (\ref{sec:ap}) is already enhancing the same benefits that the power gradient is trying to introduce.
By looking closely at the curves, however, we may note that h-ADAM is quicker for $V_3$ (higher yellow and green curves), slower for $V_4$ and working better than ADAM at $\log_{10} \eta=1$ for $V_{\rm Beale}$.
This leads to wander whether a hybrid algorithm may collect the good working regimes of both ADAM and h-ADAM in a single method.

In fact, as a last method, we introduce h(t)-ADAM, in which also the exponent $H$ becomes a function of time (see \ref{sec:ap}). The starting point is $H_0=1/2$ to exploit the power gradient at the beginning of the optimization, when conditions might be more unstable or stagnant because $V(\vec \theta_0)$ is more likely to be far from minima. Than $H_t$ converges to $1$ with the iterations $t$ to exploit the usual ADAM performances at a stage when the optimization is more refined.
In the last row of Fig.~\ref{fig:gr} we see that indeed h(t)-ADAM performs better than both ADAM and h-ADAM with fixed $H=1/2$.

\section{Conclusions}
\label{concl}
The power gradient is an intriguing modification of the standard gradient of a function, whose components are squeezed in modulus around $1$ by the application of a power with exponent $H<1$.
In fact, in general such vector is not a gradient of a function anymore.
The introduction of power gradients into GD methods in our examples almost always yields better performances for Nesterov and AMSGrad methods. ADAM algorithm instead seems not particularly taking advantage of the power gradient, perhaps due to its intrinsic ability to tame steep gradients and boost flat gradients. However, ADAM is known to carry an instability that can lead to its divergence from minima after some converging period~\cite{red18}, and AMSGrad was in fact introduced to solve this issue.
Yet, if one decides to use ADAM, the modification h(t)-ADAM should be considered, as our examples show that it achieves better performances than those of a basic ADAM.
In h(t)-ADAM, also the exponent $H$ varies.
The embedding of power gradients in other GD methods (e.g.~RMSprop~\cite{tie12}, AdaGrad~\cite{duc11}, AdaDelta~\cite{zei12}, NADAM~\cite{doz16}) will be tested in future works, together with other $H\ne 1/2$ cases.

With an idealized parabolic cost function, the basic h-GD converges always to a 2-state orbit, regardless of the learning rate. This differentiates it from the standard GD, which becomes unstable and diverges if the learning rate is above a threshold. We have described how to turn this peculiar feature of the h-GD to a new method for finding the minimum of the quadratic function. The results with more elaborated h-GD algorithms applied to two-dimensional problems have shown no significant effects due to 2-state orbits, yet understanding this feature represents another good subject for future works.

In summary, our simple examples suggest that deforming gradients can improve GD.
More studies are needed in general to determine the strength and the limitations of GD methods embodying power gradients.
It should be interesting to check how the learning rate in deep neural networks would change if h-GDs are used.
In that case, h-GDs could help surfing the flat directions encountered in a high-dimensional landscape of cost functions while avoiding too large steps along steep slopes of the function.

\paragraph{Acknowledgments}
I thank the students of the course ``Laboratory of computational physics'', degree in ``Physics of Data'' at the University of Padova, for discovering and letting me know of a typo in the description of the Nesterov algorithm in the review by \citet{meh19}.


\appendix

\section{}
\label{sec:ap}

This appendix contains the h-GD algorithms used in this work. The standard versions ($H=1$ and normal gradients) can be easily recovered from these ones or from the literature.
For the sake of simplicity, we do not write the index $(i)$ of the component. It is understood that each line refers to a given component and, e.g., by $\partial_\theta V$ we mean the $i$-th component of the gradient.

{\bf h-NAG}
\begin{align}
  g_t & = \partial_\theta V(\vec \theta_t -\gamma \vec v_{t-1}) \nonumber\\
  h_t & = \textrm{sign}(g_{t}) |g_{t}|^H  \nonumber\\
  v_t & = \gamma v_{t-1} + \eta h_t  \nonumber\\
  \theta_{t+1} & = \theta_t - v_t
\end{align}
Here $v_t$ plays the role of a (negative) velocity in the inertial dynamics of NAG. The damping parameter is set to $\gamma=0.99$, and $H=1/2$ in our simulations.

{\bf h(t)-ADAM}
\begin{align}
  H_t & = \beta_3 H_{t-1} + (1-\beta_3) \nonumber\\
  g_t & = \partial_\theta V(\vec \theta_t) \nonumber\\
  h_t & = \textrm{sign}(g_{t}) |g_{t}|^{H_t}  \nonumber\\
  m_t & = \beta_1 m_{t-1} + (1-\beta_1)h_t \nonumber\\
  s_t & = \beta_2 s_{t-1} + (1-\beta_2)h_t^2 \nonumber\\
 \hat m_t & = \frac{m_t}{1-(\beta_1)^t} \nonumber\\
 \hat s_t & = \frac{s_t}{1-(\beta_2)^t} \nonumber\\
  \theta_{t+1} & = \theta_t - \eta \frac{\hat m_t}{\sqrt{\hat s_t}+\epsilon}
\end{align}
In this work the initial value of the exponent is set to $H_0 = 1/2$, from which $H_t\to 1$ by iterating the algorithm with $\beta_3=0.999$. Of course, other starting values $H_0>0$ could be chosen too.
The simpler h-ADAM does include a constant $H$, e.g.~in this work $H=1/2$.
Other parameters $\beta_1=0.9$ and $\beta_2=0.99$ follow the typical values in the literature.
The constant $\epsilon=10^{-8}$ prevents the algorithm from exploding in case of null gradient.

{\bf h-AMSGrad}
\begin{align}
  g_t & = \partial_\theta V(\vec \theta_t) \nonumber\\
  h_t & = \textrm{sign}(g_{t}) |g_{t}|^{H}  \nonumber\\
  m_t & = \beta_1 m_{t-1} + (1-\beta_1)h_t \nonumber\\
  s_t & = \beta_2 s_{t-1} + (1-\beta_2)h_t^2 \nonumber\\
 \hat m_t & = \frac{m_t}{1-(\beta_1)^t} \nonumber\\
 \hat s_t & = \max(s_t,s_{t-1}) \nonumber\\
  \theta_{t+1} & = \theta_t - \eta \frac{\hat m_t}{\sqrt{\hat s_t}+\epsilon}
\end{align}
Parameters are as above.
Note how $\hat s_t$ is modified with respect to ADAM, with the aim of removing its potential instability.


\end{document}